\documentclass[11pt]{article}
\usepackage[dvips]{graphicx}
\usepackage{subfig}
\usepackage{amssymb}

\usepackage{color,hyperref}
\usepackage{amsmath,amscd}
\usepackage{amsthm}
\usepackage{comment}
\usepackage{breakurl}
\usepackage{mathtools, cuted}

\theoremstyle{plain}

\theoremstyle{plain}

\numberwithin{equation}{section}\graphicspath{{./figs/}}
\newcommand{\vc}[1]{\mathbf{#1}}

\newcommand{\be}{\begin{equation}}
\newcommand{\ee}{\end{equation}}

\usepackage[normalem]{ulem}

\newcounter{guocomm}
\newcounter{Note}
\definecolor{blue-violet}{rgb}{0.54, 0.17, 0.89}
\definecolor{mygreen}{rgb}{0.0, 0.5, 0.0}
\definecolor{awesome}{rgb}{1.0, 0.13, 0.32}
\definecolor{bostonuniversityred}{rgb}{0.8, 0.0, 0.0}
\newcommand\todo[1]{} % This is to turn off todo's.

\begin{document}

\title{{\bf Matrix Neural Networks}}%

\author{Junbin Gao\\
Discipline of Business Analytics\\
The University of Sydney Business School\\
Camperdown NSW 2006, Australia\\
junbin.gao@sydney.edu.au\\
\and Yi Guo\thanks{To whom future correspondences should be addressed.}\\
Centre for Research in Mathematics\\
School of Computing, Engineering and Mathematics\\
Western Sydney University\\
Parramatta NSW 2150, Australia\\
y.guo@westernsydney.edu.au\\
\and Zhiyong Wang\\
School of Information Technologies\\
The University of Sydney\\
zhiyong.wang@sydney.edu.au}%

\maketitle

\begin{abstract}
Traditional neural networks assume vectorial inputs as the network is arranged as layers of single line of computing units called neurons. This special structure requires the non-vectorial inputs such as matrices to be converted into vectors. This process can be problematic. Firstly, the spatial information among elements of the data may be lost during vectorisation. Secondly, the solution space becomes very large which demands very special treatments to the network parameters and high computational cost. To address these issues, we propose matrix neural networks (MatNet), which takes matrices directly as inputs. Each neuron senses summarised information through bilinear mapping from lower layer units in exactly the same way as the classic feed forward neural networks. Under this structure, back prorogation and gradient descent combination can be utilised to obtain network parameters efficiently. Furthermore, it can be conveniently extended for multimodal inputs. We apply MatNet to MNIST handwritten digits classification and image super resolution tasks to show its effectiveness. Without too much tweaking MatNet achieves comparable performance as the state-of-the-art methods in both tasks with considerably reduced complexity.
\end{abstract}


% ----------------------------------------------------------------
\section{Introduction}
Neural networks especially deep networks \cite{HintonOsinderoTeh2006,LeCunBengioHinton2015} have attracted a lot of attention recently due to their superior performance in several machine learning tasks such as face recognition, image understanding and language interpretation. The applications of neural netowrks go far beyond artificial intelligence domain, stretching to autonomous driving systems \cite{AngelovaKrizhevskyVanhoucke2015,KornhauserXiao2015}, pharmaceutical research \cite{UnterthinerMayrKlambauerSteijaertWegnerCeulemansHochreiter2015a,UnterthinerMayrKlambauerSteijaertWegnerCeulemansHochreiter2015}, neuroscience \cite{BernikerKording2015,DoshiKiraWagner2015,SukShenInitiativeothers2015,WangMalaveCipollini2015,YaminsCohenHongKanwisherDiCarlo2015} among others. Because of its usefulness and tremendous application potential, some open source software packages are made available for research such as caffe \cite{JiaShelhamerDonahueKarayevLongGirshickGuadarramaDarrell2014,TurchenkoLuczak2015} and Theano \cite{BergstraBreuleuxBastienLamblinPascanuDesjardinsTurianWarde-FarleyBengio2010}. Furthermore, there are even efforts to build integrated circuits for neural networks \cite{HammerstromNarayanan2015,QiaoMostafaCorradiOsswaldStefaniniSumislawskaIndiveri2015,Service2014}.

Evolving from the simplest perceptron \cite{Rosenblatt1957} to the most sophisticated deep learning neural networks \cite{LeCunBengioHinton2015}, the basic structure of the most widely used neural networks remains almost the same, i.e. hierarchical layers of computing units (called neurons) with feed forward information flow from previous layer to the next layer \cite{Bishop1995}. Although there is no restriction on how the neurons should be arranged spatially, traditionally they all line in a row or a column just like elements in a vector. The benefit of this is apparently the ease of visualisation of networks as well as the convenience of deduction of mathematical formulation of information flow. As a consequence, vectors are naturally the inputs for the neural networks. This special structure requires the non-vectorial inputs especially matrices (e.g. images) to be converted into vectors. The usual way of vectorising a matrix or multi mode tensor is simply concatenating rows or columns into a long vector if it is a matrix or flatten everything to one dimension if it is a tensor. We are mostly interested in matrices and therefore we restrict our discussion on matrices from now on. Unfortunately this process can be problematic. Firstly, the spatial information among elements of the data may be lost during vectorisation. Images especially nature images have very strong spatial correlations among pixels. Any sort of vectorisation will certainly result in the loss of such correlation. Moreover, the interpretability is heavily compromised. This renders the neural networks as ``black boxes'' as what is going on inside the network is not interpretable by human operator as the information encoded in the parameters or neurons deviates from the form we would normally percept from the very beginning if we take images as an example. Secondly, the solution space becomes very large which demands very special treatments to the network parameters. There are many adverse effects. First, the chance of reaching a meaningful local minimum is reduced due to large domain for sub-optimum. Second, the success of training relies heavily on human intervention, pretraining, special initialisation, juggling parameters of optimisation algorithms and so on. This situation becomes even worse with the growth of the depth of the networks. This is the well known model complexity against learning capacity dilemma \cite{Vapnik1995}. Third, if the spatial information among elements in matrices has to be utilised by the network, one has to resort to either specially designed connection configuration among neurons if it is possible or priors on the network parameters as regularisation which may cripple back prorogation based optimisation because spatial connection means coupling. For large scale problems e.g. big data, this may not be viable at all. Fourth, the computational cost is very high which requires massive computation platforms.

To address the issues discussed above, we propose matrix neural networks or MatNet for short, which takes matrices directly as inputs. Therefore the input layer neurons form a matrix, for example, each neuron corresponds to a pixel in a grey scale image. The upper layers are also but not limited to matrices. This is an analogy to the neurons in retina sensing visual signal which are organised in layers of matrix like formation \cite{Rodieck1973}. It is worth of pointing out that the convolutional neural network (ConvNet) \cite{LeDenkerHendersonHowardHubbardJackel1990,LeCunBottouBengioHaffner1998} works on images (matrices) directly. However, the major difference between ConvNet and MatNet is that ConvNet's input layers are feature extraction layers consisting of filtering and pooling and its core is still the traditional vector based neural network. While in MatNet matrices are passing through each layer without vectorisation at all. To achieve this, each neuron in MatNet senses summarised information through bilinear mapping from immediate previous layer units' outputs plus an offset term. Then the neuron activates complying with the pre-specified activation function e.g. sigmoid, tanh, and rectified linear unit (reLU) \cite{NairHinton2010}  to generate its output for the next layer. It is exactly the same way as the classic feed forward neural networks. Obviously the bilinear mapping is the key to preserve matrix structure. It is also the key for the application of simple back prorogation to train the network. This will become very clear after we formulate the MatNet model in the next section. In order not to disturb the flow, we leave the derivation of the gradients to appendix where interested readers can find the details.

To demonstrate the usefulness of the proposed MatNet, we will test it in two image processing tasks, the well-known MNIST handwritten digits classification and image super resolution. For digits classification, it is just a direct application MatNet to normalised images with given class labels, where MatNet acts as a classifier. However, for image super resolution, MatNet needs some adaptation, i.e. an ``add-on'' to accommodate multimodal inputs. As we will show in Section \ref{sec:mmMatNet}, this process is straightforward with great possibility to embrace other modalities such as natural languages for image understanding \cite{YangLiFermullerAloimonos2015} and automated caption generation \cite{UshikuYamaguchiMukutaHarada2015}. As shown in Section \ref{sec:exp}, MatNet can achieve comparable classification rate as those sophisticated deep learning neural networks. We need to point out that MatNet is not optimised for this task and the choices of the key network parameters such as the number of layers and neurons are somewhat arbitrary. Surprisingly for super resolution task, MatNet has superior results already in terms of peak signal to noise ratio (PSNR) compared to the state-of-the-art methods such as the sparse representation (SR) \cite{YangWrightHuangM2010}. Once again, this result can be further optimised and we will discuss some further developments that will be carried out in near future in Section \ref{sec:discussion}.

\section{Matrix Neural Network Model}\label{sec:mnnetmodel}
The basic model of a layer of MatNet is the following bilinear mapping
%\begin{align}
%\mathcal{Y} = \sigma (\mathcal{X}\times_1 U_1 \times_2 \cdots \times_N U_N + \mathcal{B}) + \mathcal{E}. \label{eq:mnnetmodel}
%\end{align}
%where $\mathcal{X}$ and $\mathcal{Y}$ are the input and output tensors of this layer respectively. In \eqref{eq:mnnetmodel}, $U_i$ for $i=1,\ldots,N$, is the weights on the connection from the previous layer, $\times_i$ is the mode $i$ multiplication of the tensor, $\mathcal{B}$ is the offset of current layer, $\sigma()$ is the activation function acting on each element of the tensor and $\mathcal{E}$ is the error. Note that we use tensor for generality. For the sake of simplicity, we consider the case of $N=2$, i.e., both $\mathcal{X}$ and $\mathcal{Y}$ are matrix variables written as $X$ and $Y$ and \eqref{eq:mnnetmodel} becomes the following bilinear mapping
\begin{align}
Y = \sigma (UXV^T+B) + E, \label{eq:mnnetmodelmat}
\end{align}
where $U$, $V$, $B$ and $E$ are matrices with compatible dimensions, $U$ and $V$ are connection weights, $B$ is the offset of current layer, $\sigma(\cdot)$ is the activation function acting on each element of matrix and $E$ is the error.
%The purpose is to work out the backpropagation algorithm.
\subsection{Network Structure}
The MatNet consists multiple layers of neurons in the form of \eqref{eq:mnnetmodelmat}. Let $X^{(l)}\in\mathbb{R}^{I_l\times J_l}$ be the matrix variable at layer $l$ where $l = 1, 2, \ldots, L, L+1$. Layer 1 is the input layer that takes matrices input directly and Layer $L+1$ is the output layer. All the other layers are hidden layers. Layer $l$ is connected to Layer $l+1$ by
\begin{align}
X^{(l+1)} = \sigma ( U^{(l)} X^{(l)} V^{(l)T}  + B^{(l)}). \label{eq:MatNet-2}
\end{align}
where $B^{(l)}\in\mathbb{R}^{I_{l+1}\times J_{l+1}}$, $U^{(l)}\in\mathbb{R}^{I_{l+1}\times I_l}$ and $V^{(l)}\in\mathbb{R}^{J_{l+1}\times J_l}$, for $l = 1, 2, ..., L-1$.  For the convenience of explanation, we define
\begin{align}
N^{(l)} = U^{(l)} X^{(l)} V^{(l)T}  + B^{(l)}  \label{eq:MatNet-3}
\end{align}
for $l=1, 2, ..., L$.  Hence \[ X^{(l+1)} = \sigma (N^{(l)}).
\]
The shape of the output layer is determined by the functionality of the network, i.e. regression or classification, which in turn determines the connections from Layer $L$. We discuss in the following three cases.
\begin{itemize}
\item Case 1: Normal regression network. The output layer is actually a matrix variable as $O = X^{(L+1)}$. The connection between layer $L$ and the output layer is defined as \eqref{eq:MatNet-2} with $l = L$.
\item Case 2: Classification network I. The output layer is a multiple label (0-1) vector $\mathbf o = (o_1, ..., o_K) $ where $K$ is the number of classes. In $\mathbf o$, all elements are 0 but one 1. The final connection is then defined by
    \begin{align}
    o_k = \frac{\exp(\mathbf u_k  X^{(L)} \mathbf v^T_k  + tb_k)}{\sum^K_{k'=1} \exp(\mathbf u_{k'}  X^{(L)} \mathbf v^T_{k'}  + tb_{k'})}, \label{eq:MatNet-4}
    \end{align}
    where $k=1,2,...,K$, $\overline{U} = [\mathbf u^T_1, ...., \mathbf u^T_K]^T \in\mathbb{R}^{K\times I_L}$ and $\overline{V} = [\mathbf v^T_1, ...., \mathbf v^T_K]^T\in\mathbb{R}^{K\times J_L}$. That is both $\mathbf u_k$ and $\mathbf v_k$ are rows of matrices $\overline{U}$ and $\overline{V}$, respectively.  Similar to \eqref{eq:MatNet-3}, we denote
    \begin{align}
    n_k = \mathbf u_k  X^{(L)} \mathbf v^T_k  + tb_k.\label{eq:MatNet-5}
    \end{align}
    \eqref{eq:MatNet-4} is the softmax that is frequently used in logistic regression \cite{HosmerLemeshowSturdivant2013}. Note that in \eqref{eq:MatNet-4}, the matrix form is maintained. However, one can flatten the matrix for the output layer leading to the third case.
\item Case 3: Classification network II.  The connection of Layer $L$ to the output layer can be defined as the following
    \begin{align}
    &N^{(L)}_{k} = \text{vec}(X^{(L)})^T \overline{\mathbf u}_k + tb_k \label{eq:MatNet-6}\\
    &o_{k} = \frac{\exp(N^{(L)}_{k})}{\sum^K_{k'=1}\exp(N^{(L)}_{k'})}  \label{eq:MatNet-7}
    \end{align}
    where vec$()$ is the vectorisation operation on matrix and $\overline{\mathbf u}_k$ is a column vector with compatible length. This makes Case 2 a special case of Case 3.
\end{itemize}
%The model in Case 2 is a sub-model in Case 3.

Assume that we are given a training dataset $\mathcal{D} = \{(X_n, Y_n)\}^N_{n=1}$ for regression or $\mathcal{D} = \{(X_n, \mathbf t_n)\}^N_{n=1}$ for classification problems respectively. Then we define the following loss functions
\begin{itemize}
\item Case 1: Regression problem's loss function is defined as
\begin{align}
L = \frac1N\sum^N_{n=1}\frac12 \| Y_n - X^{(L+1)}_n\|^2_F. \label{eq:MatNet-8}
\end{align}
\item Cases 2\&3: Classification problem's cross entropy loss function is defined as
\begin{align}
L = -\frac1N\sum^N_{n=1} \sum^K_{k=1} t_{nk} \log (o_{nk}). \label{eq:MatNet-9}
\end{align}
\end{itemize}
Note that the selection of cost function is mainly from the consideration of the convenience of implementation. Actually, MatNet is open to any other cost functions as long as the gradient with respect to unknown variables can be easily obtained.
%  in deriving derivative formulas, we ignore this $\frac1N$.

%\todo{Check the following remarks carefully (from below to the next subsection)!}
From Eq. \eqref{eq:MatNet-2} we can see that the matrix form is well preserved in the information passing right from the input layer. By choosing the shape of $U^{(l)}$,  $V^{(l)}$ and $B^{(l)}$ accordingly, one can reshape the matrices in hidden layers. In traditional neural networks with vectors input, Eq. \eqref{eq:MatNet-2} actually becomes
\be\label{eq:vnnet1}
\vc x^{(2)} = \sigma(W^{(1)}\text{vec}(X^{(1)}) + \vc b^{(1)})
\ee
where $\vc x^{(2)}$ and $\vc b^{(1)}$ are column vectors with compatible lengths. If we vectorise the first hidden layer of MatNet we obtain
\be\label{eq:vectorisedMatNet}
\text{vec}(X^{(2)}) = \sigma(({V^{(1)}}^\top\otimes U^{(1)})\text{vec}(X^{(1)}) + \text{vec}(B^{(1)})),
\ee
where $A\otimes B$ is the Kronecker product between matrix $A$ and $B$ and we used the identity
\[
\text{vec}(A X B) = (B^T\otimes A)\text{vec}(X).
\]

It is clear that by choosing $W^{(1)}$ in traditional neural networks such that $W^{(1)} = {V^{(1)}}^\top\otimes U^{(1)}$, it is possible to mimic MatNet and it is also true for other layers. Therefore, MatNet is a special case of traditional neural networks. However, ${V^{(l)}}^\top\otimes U^{(l)}$ has significantly less degrees of freedom than $W^{(l)}$, i.e. $I_{l+1}I_l + J_{l+1}J_l$ v.s. $I_{l+1}I_lJ_{l+1}J_l$. The reduction of the solution space brought by the bilinear mapping in Eq. \eqref{eq:MatNet-2} is apparent. The resultant effects and advantages include less costly training process, less local minima, easier to handle and most of all, direct and intuitive interpretation. The first three comes immediately from the shrunk solution space. The improved interpretability comes from the fact that $U^{(l)}$ and $V^{(l)}$ work on the matrices directly which normally correspond to input images. Therefore, the functions of $U^{(l)}$ and $V^{(l)}$ becomes clearer, i.e. the linear transformation applied on matrices. This certainly connects MatNet to matrix or tensor factorisation type of algorithms such as principal component analysis \cite{HopkinsShiSteurer2015,PaateroTapper1994,ZhouCichockiXie2012} broadening the understanding of MatNet.

%One remarkable feature of MatNet is that its bilinear form of information passing stems from the more general tensor form defined in Eq. \eqref{eq:MatNetmodel} where a multilinear relation is stated. Furthermore, the way the information passing and similar derivation of gradient of multilinear form to bilinear mapping suggest that MatNet can be ``upgraded'' to networks sensing tensors. All the apparatus of MatNet such as BP, gradient descent, as well as the priors and regularisation on weights can be readily used. And indeed we reserve this as our future research.

\subsection{Optimisation}
We collect all the unknown variables i.e. the network parameters of each layer here. They are $U^{(l)}$, $V^{(l)}$, $B^{(l)}$ for $l=1,\ldots,L$, and $\overline{\mathbf u}_k$ and $tb_k$ for the output layer. Write the parameters of each layer as $\Theta^{(l)}$. From Eq. \eqref{eq:MatNet-2} one can easily see that the information is passing in the exactly the same way of the traditional fee forward neural networks. The underlining mechanism is the bilinear mapping in \eqref{eq:MatNet-3}, which preserves the matrix form throughout the network. This suggests that the optimisation used in traditional neural networks, i.e. back propagation (BP) and gradient descent combination can be used for MatNet. All we need to do is to obtain the derivative of the cost function w.r.t $\Theta^{(l)}$, which can be passed backwards the network.

Since we proposed both regression and classification network models, the derivatives differ slightly in these two cases due to different cost functions while the back propagation is exactly the same. The details about the gradients and back propagation are in the appendix for better flow of the paper. Once the gradients are computed, then any gradient descent algorithm such as the limited memory Broyden-Fletcher-Goldfarb-Shanno (LBFGS) \cite{GaoReynoldsothers2004} can be readily used to find the sub-optimum given an initialisation. Normally, the network is initialised by random numbers to break symmetry. When the number of layers of a MatNet is 3, this strategy is good enough. However, if MatNet contains many layers, i.e. forming a deep network, then the complexity of the model increases drastically. It requires more training samples. Meanwhile some constraints will be helpful for faster convergence or better solution.

\subsection{Regularisation}
Although MatNet has reduced solution space heavily by using the bilinear mapping in \eqref{eq:MatNet-3} already, some techniques routinely used in traditional neural networks can still be used to further constrain the solution towards the desired pattern. The first is the weight decay, i.e. clamping the size of the weights on the connections, mainly $U^{(l)}$ and $V^{(l)}$. Normally we use Frobenius norm of a matrix for this purpose, that is to incorporate
\[
\lambda \sum_l(\|U^{(l)}\|_F^2+\|V^{(l)}\|_F^2),
\]
where $\lambda$ is a nonnegative regularisation parameter and the summation of Frobenius norms includes the output layer as well.

One may immediately think of the sparsity constraint on the weights to cut off some connections between layers similar to the DropConnect in \cite{WanZeilerZhangCunFergus2013}. It turns out that it is not trivial to incorporate sparsity constraint manifested by sparsity encouraging norms such as $\ell_1$ norm favourably used in sparse regressions \cite{Tibshirani1996}. The dropping in \cite{WanZeilerZhangCunFergus2013} in implemented by a 0/1 mask sampled from Bernoulli distribution. Here we discuss another type of sparsity which is much easier to be incorporated into MatNet. This is the situation when we have an over supply of neurons in hidden layers. In this case, the neural network may be able to discover interesting structure in the data with less number of neurons.

Recall that $X^{(l)}_n$ in \eqref{eq:MatNet-2}  denotes the activation at hidden unit $l$ in the network. let
\begin{align}
\overline{\rho}^{(l)} = \frac1N\sum^N_{n=1} X^{(l)}_n \end{align} be the average activations of hidden layer $l$ (averaged over the training set).  %Note: in deriving derivative formulas, we ignore this $\frac1N$.
Through (approximately) enforcing the constraint elementwise %\todo{[{\color{red}Please note $\overline{\rho}^{(l)}$ is a matrix. - Junbin}]}
\[
\overline{\rho}^{(l)}_{ij} = \rho,
\]
one can achieve sparsity in reducing the number of neurons \cite{ShuFyshe2013}. Therefore, $\rho$ is called a sparsity parameter, typically a small value close to zero, e.g. $\rho = 0.05$. In words, the constraint requires the average activation of each hidden neuron  to be close to a small given value. To satisfy this constraint, some hidden units' activations must  be close to 0.

To implement the above equality constraint, we need a penalty term  penalising the elements of $\overline{\rho}^{(l)}$ deviating significantly from $\rho$.  The deviation is quantified as the following akin to Kullback-Leibler divergence or entropy \cite{CoverThomas2006}:
\begin{align}
R_l = \text{sum}\left( \rho \log\frac{\rho}{\overline{\rho}^{(l)}} + (1-\rho)\log\frac{1-\rho}{1-\overline{\rho}^{(l)}}\right) \label{eq:rl}
\end{align}
where $\text{sum}(M)$ summing over all the elements in matrix $M$; $\log$ and $/$ are applied to matrix elementwise. %The reason for choosing \eqref{eq:rl} is due to its effectiveness and its simple form of gradient.
To screen out neurons that are not necessary, we add the following extra term in the cost function of MatNet
\[
\beta\sum^{L}_{l=2} R_l.
\]
The gradient of this term is detailed in the appendix.

\begin{comment}
Other regularisations can be considered in MatNet. In particular we want to mention the manifold constraint on connection weights proposed in  \cite{BadrinarayananMishraCipolla2015} proven to be useful when extending MatNet to deep architectures, i.e. with many hidden layers which may include max-pooling and sub-sampling layers. The idea is to restrict the connection weights, i.e. $U^{(l)}$ and $V^{(l)}$ in MatNet, to be always on unit-norm manifold. This constraint helps improve accuracy of the network effectively.
\end{comment}

\section{Multimodal Matrix Neural Networks}\label{sec:mmMatNet}
We have the basics of MatNet from above discussion. Now we proceed to extending MatNet to multimodal case for image super resolution application. The extension is as straightforward as including more than one input matrix at the same time at input layer. Conceptually, we have more than one input layer standing side by side for different modalities and they all send the information to the shared hidden layers through separate connections \cite{NgiamKhoslaKimNamLeeNg2011}. It turns out for super resolution, three layer MatNet is sufficient, i.e., input layer, hidden layer and output layer, and it works on autoencoder \cite{HintonSalakhutdinov2006} mode meaning a regression MatNet reproducing the input in output layer. This requires that the output layer has the same amount of modalities as the input layer. Although we showcase only a three layer regression multimodal MatNet, it is not difficult to extend to other type of multimodal MatNet with multiple hidden layers using the same methodology.

%\subsection{Network Structure}
Assume $D$ modalities as matrices in consideration denoted by $X^j \in \mathbb{R}^{K_{j1}\times K_{j2}}$ ($j=1,2,...,D$). Similarly there are $D$ output matrix variables of the same sizes. Denote by $\mathcal{X} = (X^1, ..., X^D)$. In the hidden layer, we only have one matrix variable $H \in \mathbb{R}^{K_1\times K_2}$. The transformation from input layer to hidden layer is defined by the following multiple bilinear mapping with the activation function $\sigma$ (sigmoid or any other activation function)
\begin{align}
H = \sigma(\sum^D_{j=1} U_j X^j V^T_j + B) \label{eq:MatNetmm-1}
\end{align}
and from hidden layer to output layer by
\begin{align}
\widehat{X}^j = \sigma(R_j H S^T_j + C_j), \;\; j = 1, 2, ..., D. \label{eq:MatNetmm-2}
\end{align}

We call $H$ the encoder for data $\mathcal{X}$. For a given set of training data $\mathcal{D} = \{\mathcal{X}_j\}^N_{i=1}$ with $\mathcal{X}_i = (X^1_i, ..., X^D_i)$,  the corresponding hidden variable is denoted by $H_i$.  The objective function to be minimised for training an MatNet autoencoder is defined by
\begin{align}
L = \frac1{2N}\sum^N_{i=1} \sum^D_{j=1}\| \widehat{X}^j_i -  X^j_i\|^2_F.  \label{eq:MatNetmm-3}
\end{align}
$L$ is a function of all the parameters $W = \{U_j, V_j, R_j, S_j, C_j, B\}^D_{j=1}$.

We leave the derivation of the gradients of multimodal MatNet autoencoder to the appendix. It is very similar to those of the original MatNet and therefore the the same BP scheme can be utilised for optimisation.

\section{Experimental Evaluation}\label{sec:exp}
In this section, we apply MatNet to MNIST handwritten digits classification and image super resolution. The network settings are somewhat arbitrary, or in other words, we did not optimise the number of layers and neurons in each layer in these tests. For handwritten digits recognition, MatNet was configured as a classification network, i.e. the output layer was a vector of softmax functions as in Eq. \eqref{eq:MatNet-6} and \eqref{eq:MatNet-7} of length 10 (for 10 digits). For illustration purpose, we selected a simple MatNet. It contained 2 hidden layers, each with $20\times20$ and $16\times16$ neurons. As the numbers of layers and neurons were very conservative, we turned off sparsity constraint as well as  %MatNet was onboarded with unit norm manifold constraint. %\todo{\underline{We on boarded with unit norm manifold constraint.}[{\color{red}Please check it. Yes we did for MNIST.}]} Hence we excluded the 
weights decay. For super resolution task, the only hidden layer was of size $10\times10$, therefore, only 3 layer MatNet. %\todo{\underline{Therefore, only 3 layer MatNet.} Yes, it is so}
The activation function in both networks was sigmoid.

\subsection{MNIST Handwritten Digits Classification}
\begin{figure}
  \centering
  \includegraphics[width=0.9\linewidth]{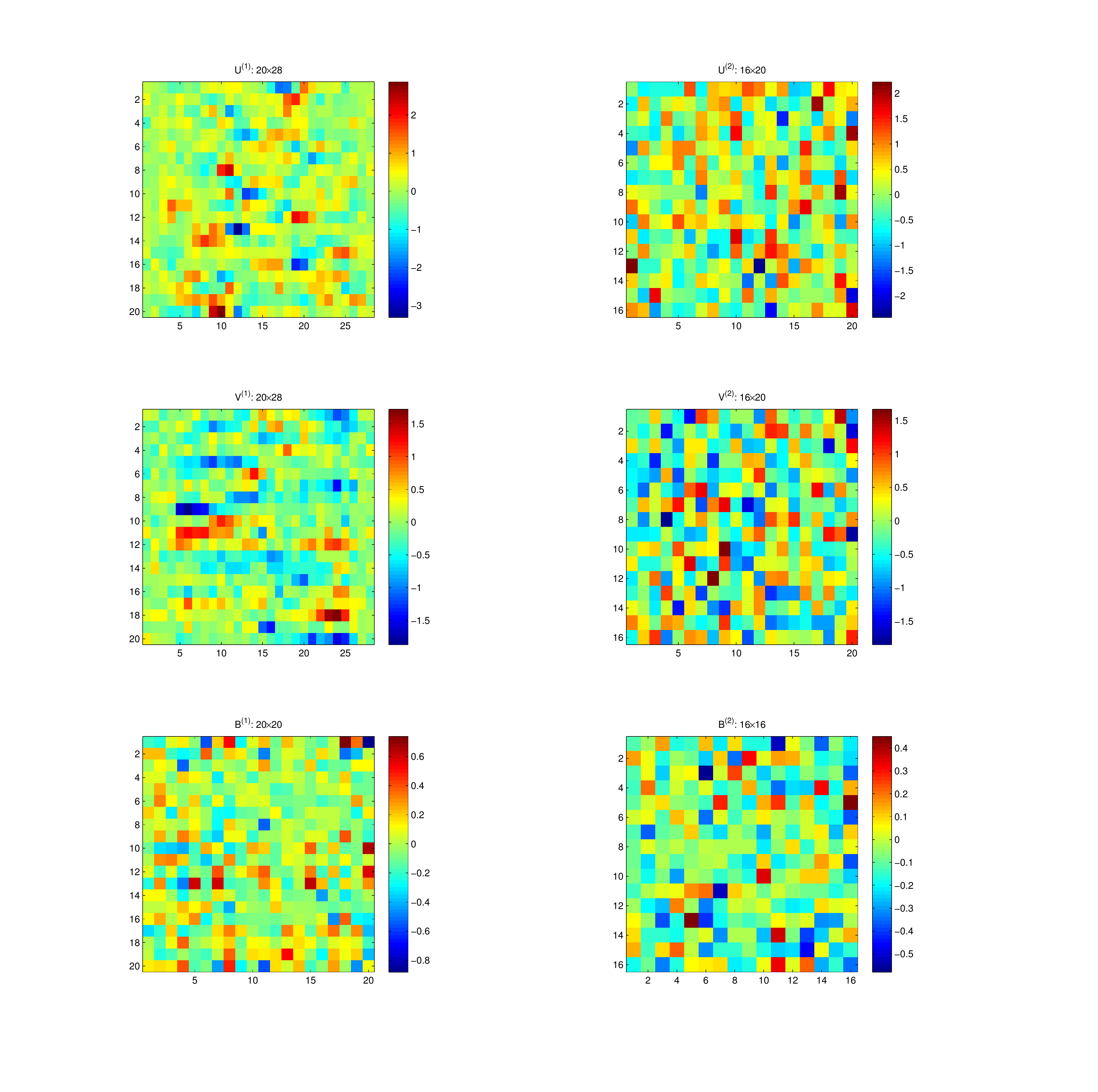}\\
  \caption{Weights and bias learnt by MatNet classifier.}\label{fig:weightsmnist}
\end{figure}

\begin{figure}
\centering
  \subfloat[First 100 test digits.]{\includegraphics[width=0.3\linewidth]{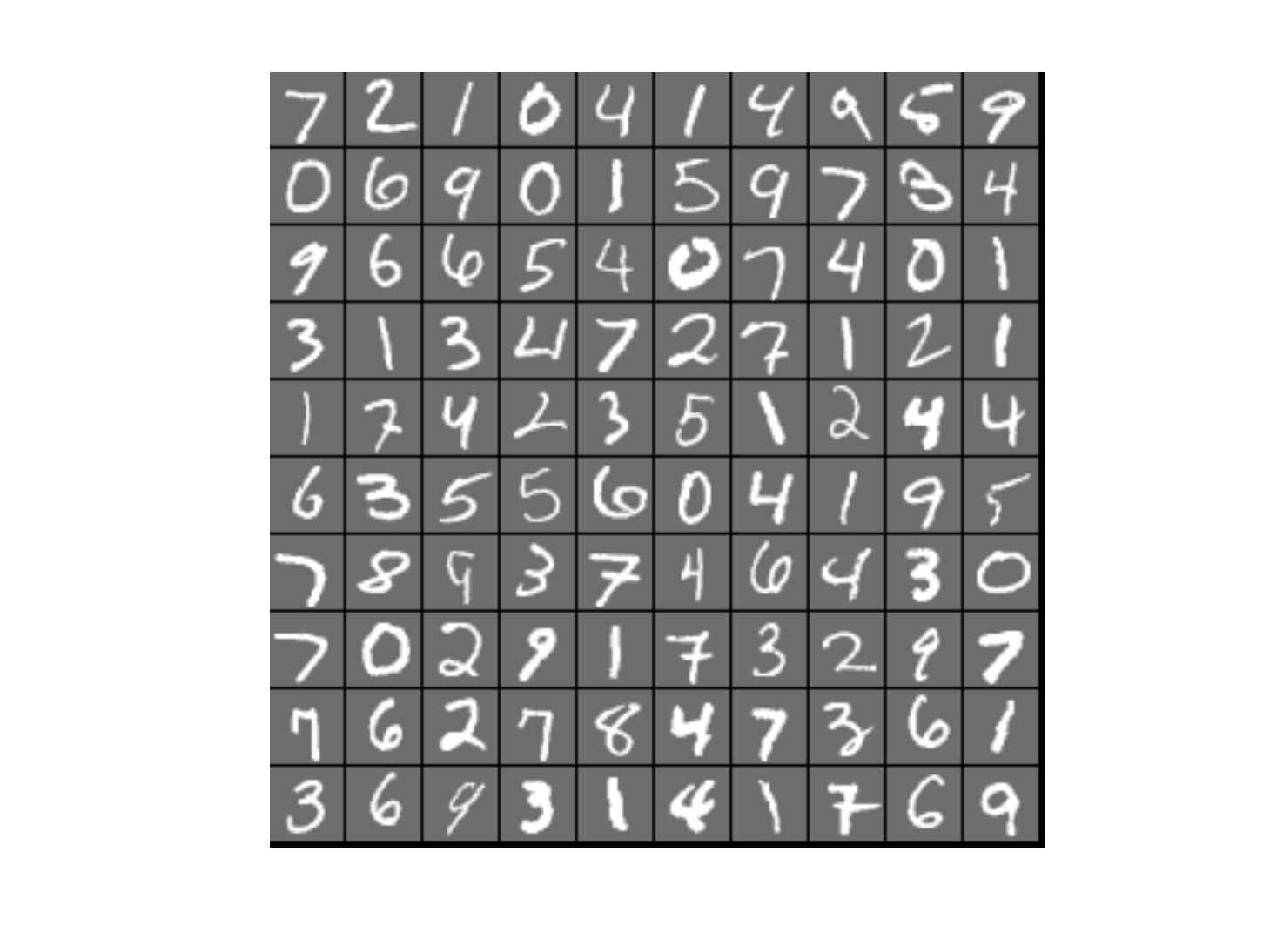}}\label{fig:mnist100testdigits}
  \subfloat[Hidden layer 1 output.]{\includegraphics[width=0.3\linewidth]{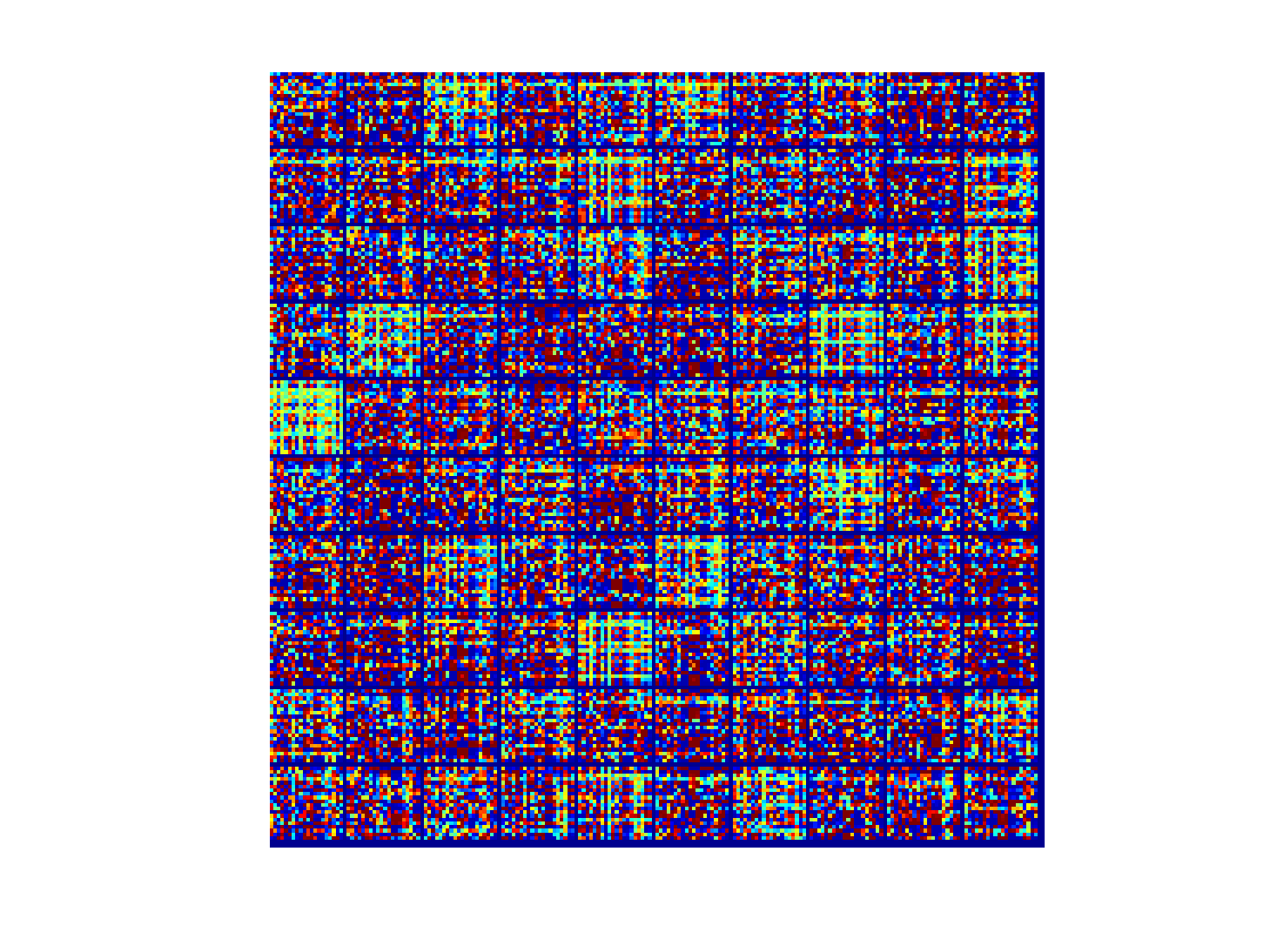}}\label{fig:hl1mnist100testdigits}
  \subfloat[Hidden layer 2 output.]{\includegraphics[width=0.3\linewidth]{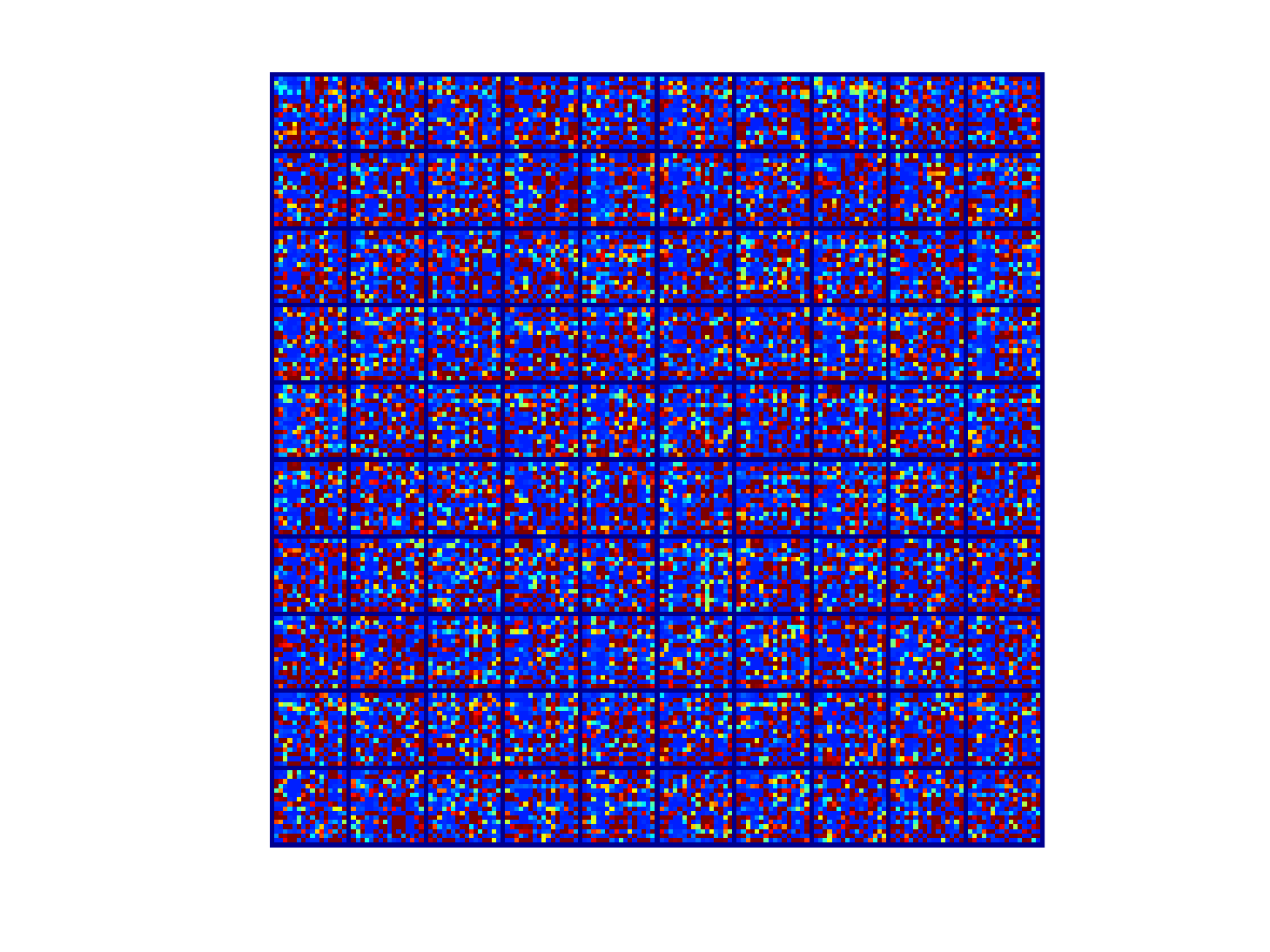}}\label{fig:hl2mnist100testdigits}

\caption{Hidden layer output of MatNet for MNIST dataset. }\label{fig:MatNetforMNISTHLoutput}
\end{figure}

The MNIST handwritten digits database is available at \url{http://yann.lecun.com/exdb/mnist/}. The entire database contains 60,000 training samples and 10,000 testing samples, and each digit is a $28\times 28$ gray scale image. We use all training samples for modeling and test on all testing samples. Figure \ref{fig:weightsmnist} shows the weights, $U^{(l)}$ and $V^{(l)}$, and bias $B^{(l)}$ in hidden layers. %\todo{[{\color{red}Better to use grayscale images. - Junbin}] I tried grayscale first, but they don't look nice. }
Figure \ref{fig:MatNetforMNISTHLoutput} shows the first 100 test digits, and hidden layer outputs. The check board effects can be seen from the the hidden layer output in Figure \ref{fig:MatNetforMNISTHLoutput}(b). The final test accuracy is 97.3\%, i.e. error rate of 2.7\%, which is %\todo{Maybe we cross out ``slightly''?}
inferior to the best MNIST performance by DropConnect with error rate 0.21\%.  

\begin{figure}
  \centering
  \includegraphics[width=0.8\linewidth]{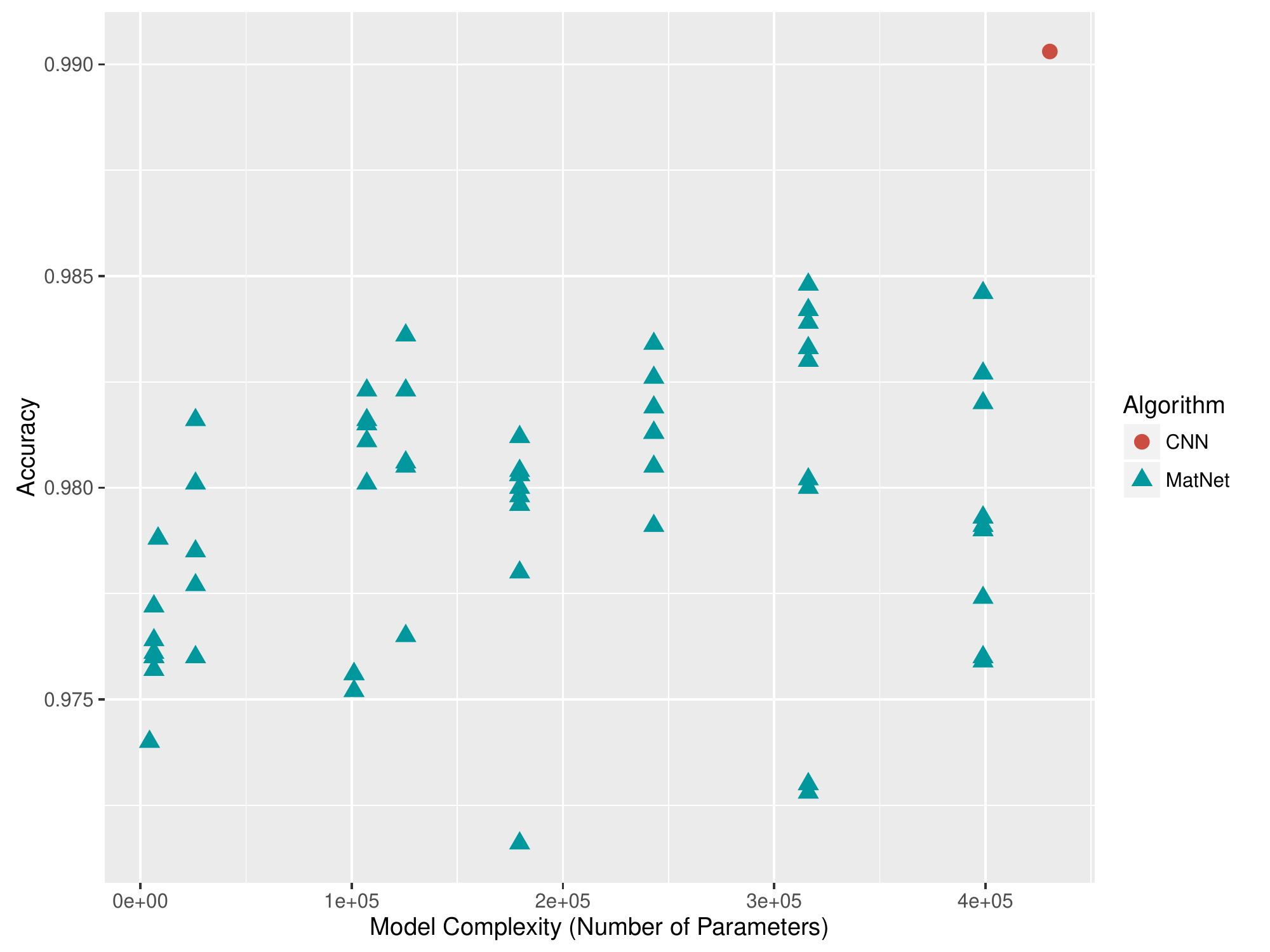}\\
  \caption{Test accuracy of MatNet vs CNN}\label{fig:matnetvscnn}
\end{figure}

\begin{figure*}
  \centering
  \includegraphics[width=0.8\linewidth]{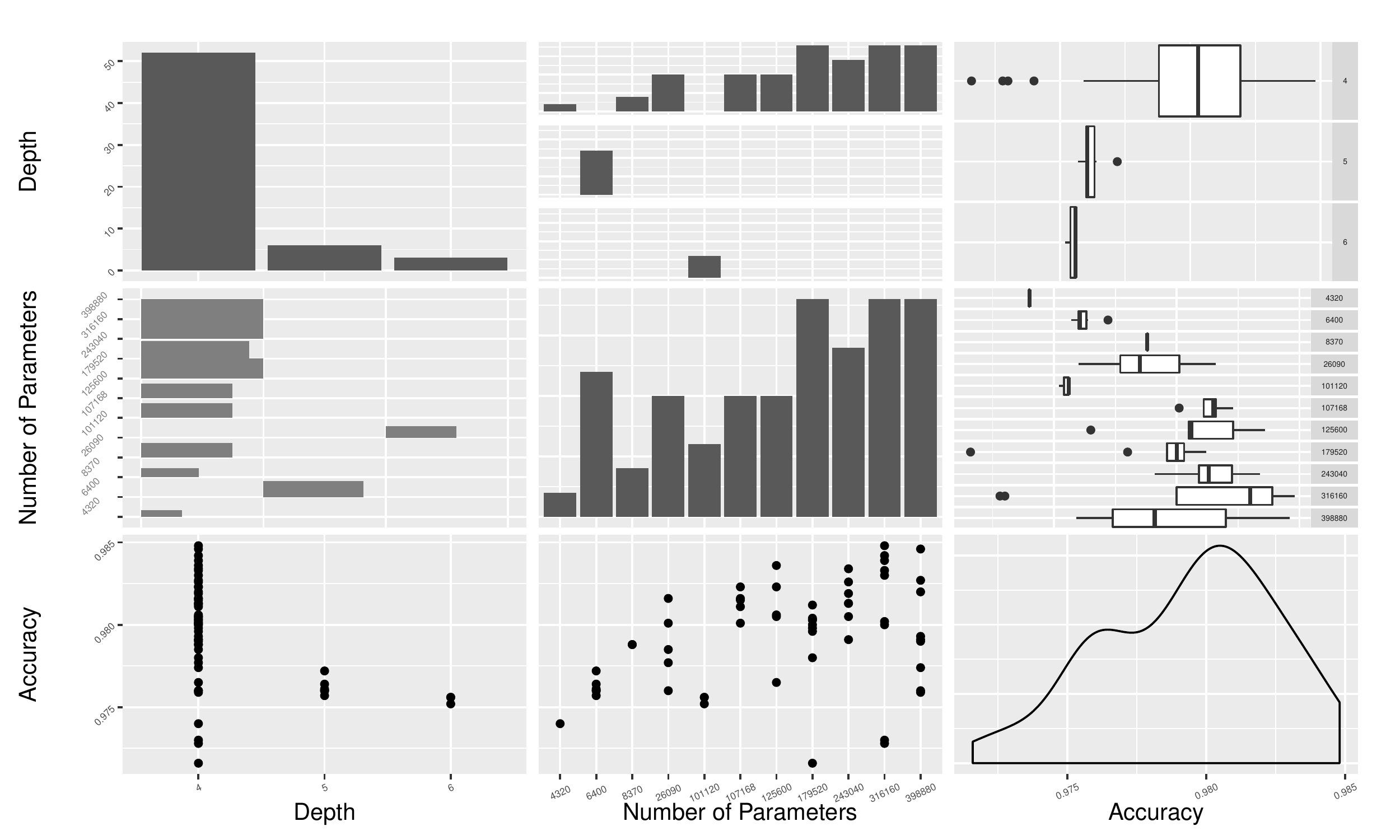}\\
  \caption{Some statistics of MatNet in this experiment. }\label{fig:matnetmulti}
\end{figure*}

However, as we stated earlier, MatNet has much less computational complexity. To see this clearly, we carried out a comparison between MatNet and ``plain'' convolutional neural networks (CNN), i.e. CNN without all sorts of ``add-ons''. The CNN consisted of two convolutional layers of size $20\times1\times5\times5$ and $50\times20\times5\times5$ one of which is followed by a $2\times2$ max pooling, and then a hidden layer of 500 and output layer of 10, fully connected. This is the structure used in Theano \cite{Al-RfouAlainAlmahairiEtAl2016} demo. The total number of parameters to optimise is 430500, while the total number of parameters in MatNet is 5536. The server runs a 6-core i7 3.3GHz CPU with 64GB memory and a NVIDIA Tesla K40 GPU card with 12GB memory. We used Theano for CNN which fully utilised GPU. On contrast, MatNet is implemented with Matlab without using any parallel computing techniques. The difference of training time is astounding. It costed the server more than 20 hours for CNN with final test accuracy of 99.07\%, whereas less than 2 hours for MatNet with test accuracy of 97.3\%, i.e. 1.77\% worse. In order to see if MatNet can approach this CNN's performance in terms of accuracy, we varied the structure of MatNet in both number of neurons in each layer and number of layers (depth). However, we limited the depth to the maximum of 6 as we did not consider deep structure for the time being. Due to the randomness of the stochastic gradient descent employed in MatNet, we ran through one structure multiple times and collected the test accuracy. Fig. \ref{fig:matnetvscnn} shows the performance of different MatNet compared against CNN. The model complexity is rendered as the number of parameters in the model, which is the horizontal axis in the plot. So when MatNet gets more complex, it approaches CNN steadily. Fig. \ref{fig:matnetmulti} shows some statistics of all the tested MatNets where the depth is also included. The bar plots are mainly histograms of given pair of variables. The diagonal panels are density for corresponding variables such as the right bottom one is the test accuracy density where it show the majority of MatNets achieved more than 98\% accuracy. The left two bottom panels show the scatter plots of accuracy against depth and number of parameters. However, the two panels on the top right summarise these as box plots are more informative. They show that the most complex models are not necessarily the best models on average. The best model (with highest test accuracy) is the one with depth of 4, i.e. two hidden layers, $160\times160$ neurons each and 316160 parameters in total that achieved 98.48\% accuracy, very close to that of CNN despite the fact that MatNet is not at all optimised in almost every aspect such as  optimisation strategy. This implies that MatNet has the potential to match the performance of CNN with more future efforts with foreseeable great savings in computation.
 
%\begin{figure}
%\centering
%\begin{minipage}{0.3\linewidth}
%  \centering
%  \includegraphics[width=1\linewidth]{mnist100testdigits}
%  \caption{First 100 test digits.}\label{fig:mnist100testdigits}
%\end{minipage}
%\begin{minipage}{0.3\linewidth}
%  \centering
%  \includegraphics[width=1\linewidth]{hl1mnist100testdigits}
%  \caption{Hidden layer 1 output. }\label{fig:hl1mnist100testdigits}
%\end{minipage}
%\begin{minipage}{0.3\linewidth}
%  \centering
%  \includegraphics[width=1\linewidth]{hl2mnist100testdigits}
%  \caption{Hidden layer 2 output.}\label{fig:hl2mnist100testdigits}
%\end{minipage}
%%\caption{Hidden layer output of MatNet for MNIST dataset. }\label{fig:MatNetforMNISTHLoutput}
%\end{figure}

\subsection{Image Super Resolution}
For image super resolution, we need to use the multimodal MatNet detailed in Section \ref{sec:mmMatNet}.  The training is the following. From a set of high resolution images, we downsample them by bicubic interpolation to the ratio of $1/s$ where $s$ is the target up-scaling factor. In this experiment, $s=2$. From these down scaled images, we sampled patches, say $15\time15$, from their feature images, i.e. first and second derivatives along x and y direction, 4 feature images for each. These are the modalities from $X^2$ to $X^5$. We also sampled the same size patches from the original high resolution images as $X^1$. See Eq. \eqref{eq:MatNetmm-1}. These data were fed into multimodal MatNet for training. %Note that if the images are full color, we converted them to gray scale first.

To obtain a high resolution image we used the following procedure. First upscale the image by bicubic interpolation to the ratio of $s$ and convert it to YCbCr space. The luminance component is then the working image on which the same size patches are sampled by sliding window as new input $X^1$. Obtain 4 feature images from this working image on which patches are sampled exactly the same way to form $X^2$ to $X^5$. Feed these to a well trained multimodal MatNet to get high resolution image patches from network output. The high resolution patches are then merged together by averaging pixels in patches. This gives us the high resolution luminance image, which is in turn combined with up-scaled, chrominance images, Cb and Cr images, simply by bicubic interpolation, to form final high resolution image in YCbCr space. For better display, it is converted to RGB format as final image.

\begin{figure}
  \centering
  \includegraphics[width=1\linewidth]{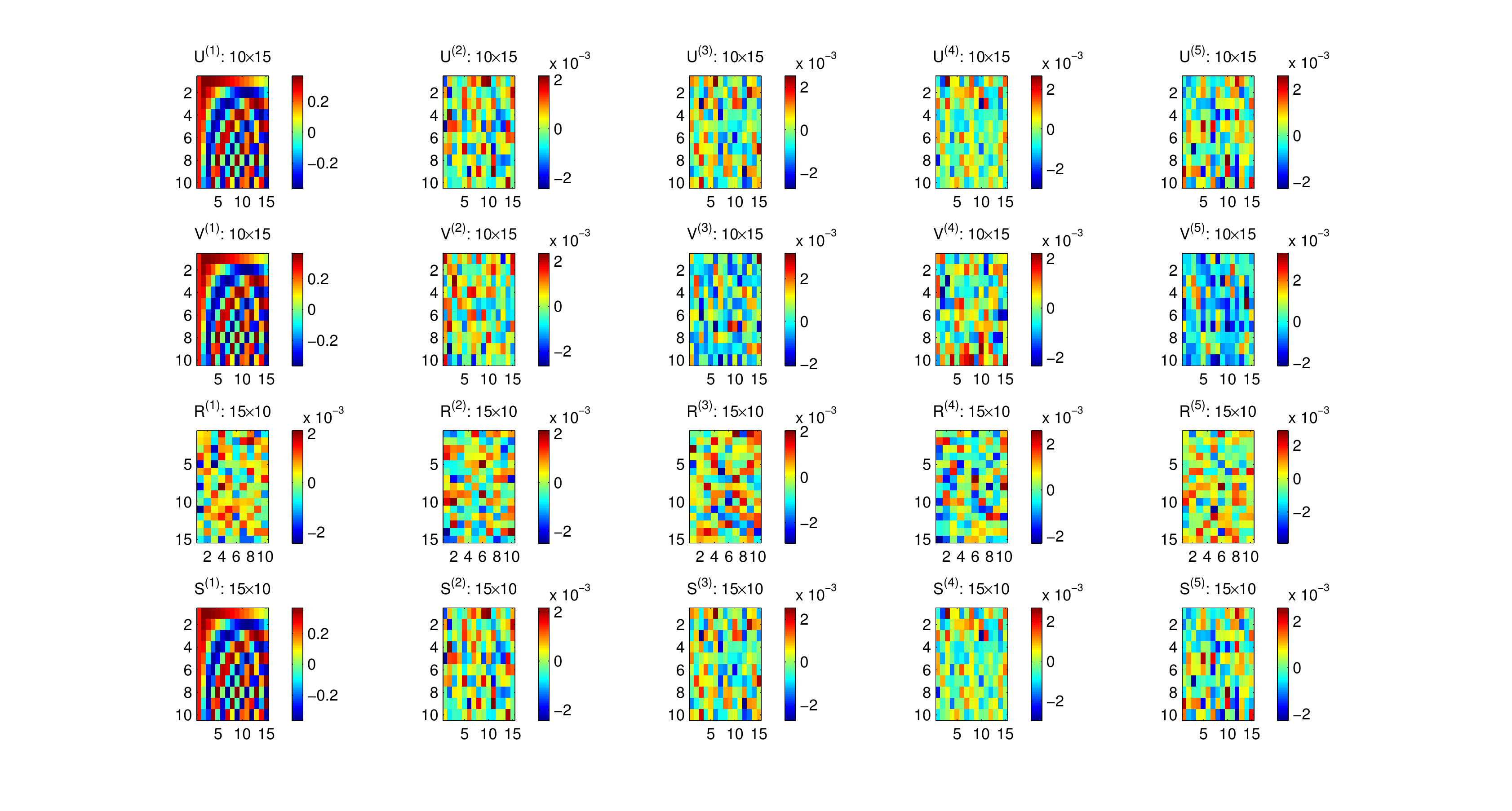}\\
  \caption{Multimodal MatNet weights learnt for super resolution.}\label{fig:srweights}
\end{figure}

We applied MatNet to the data set used in SR \cite{YangWrightHuangM2010}, both for training and testing. There are 69 images for training. The patch size was $15\times15$. We randomly sampled 10,000 patches altogether from all images for training. Some additional parameters for MatNet are $\lambda = 0.001$, $\rho = 0.05$ and $\beta=1$. So we turned on weight decay and sparsity constraints but left out the manifold constraint. Figure \ref{fig:srweights} shows the network parameters learnt from the data, from which we can observe the scaling changing filters in the weights for high resolution patches.

\begin{figure}
  \centering
  \subfloat[Lena image ($128\times128$)]{\includegraphics[width=0.9\linewidth]{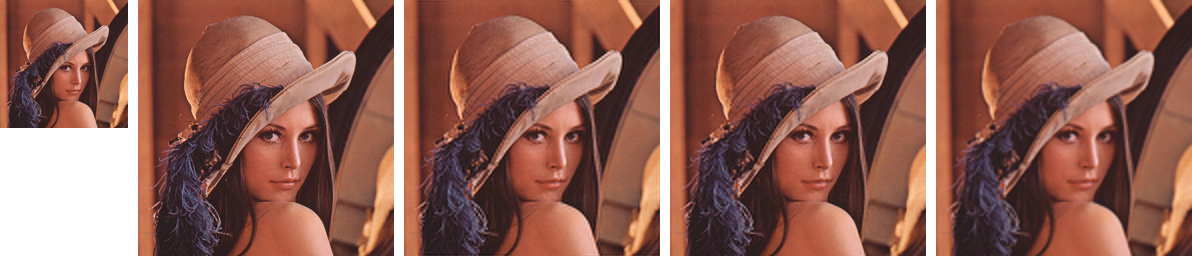}}\\
  \subfloat[Kingfisher image ($256\times256$)]{\includegraphics[width=0.9\linewidth]{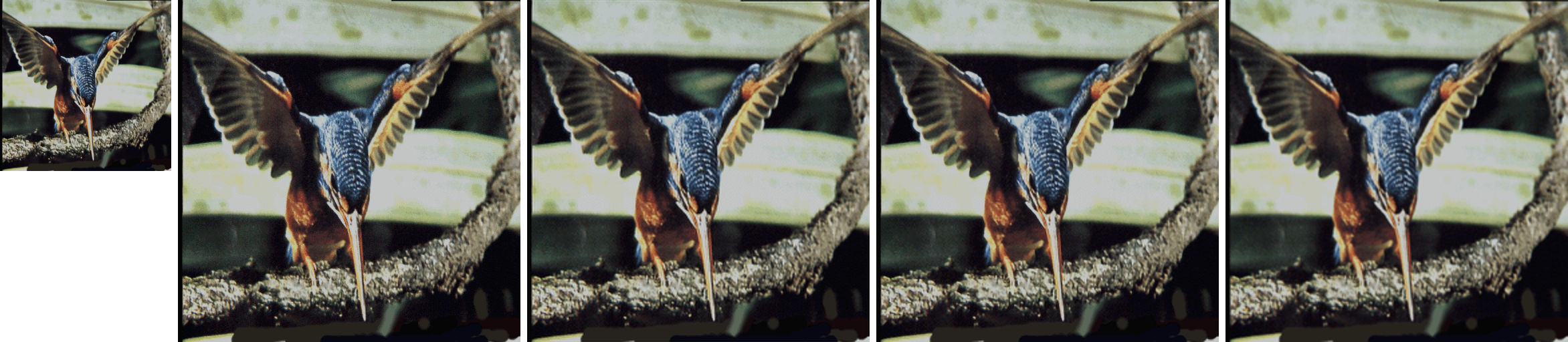}}
  \caption{Super resolution on 2 sets of testing images. From left to right: input small size image, true high resolution image, up-scaled images (2 times) produced by multimodal MatNet, SR and bicubic interpolation respectively. }\label{fig:srimages}
\end{figure}

\begin{figure}
  \centering
  \subfloat[All 12 test images]{\includegraphics[width=0.45\linewidth]{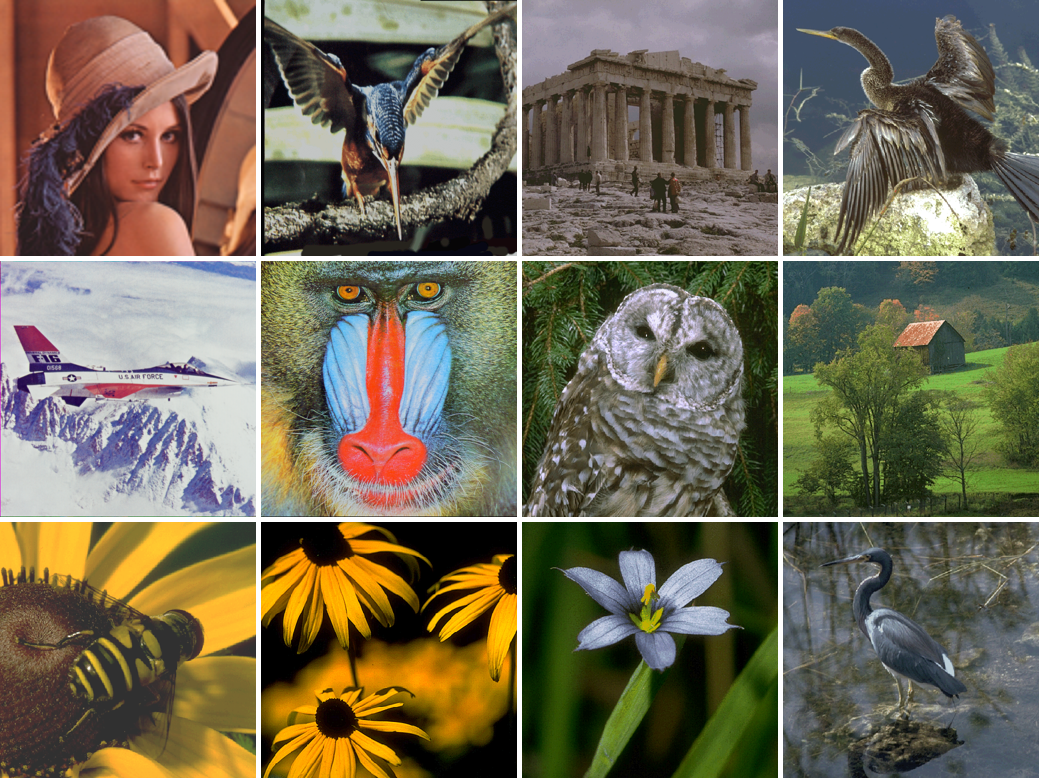}}
  \subfloat[PSNR results]{\includegraphics[width=0.45\linewidth]{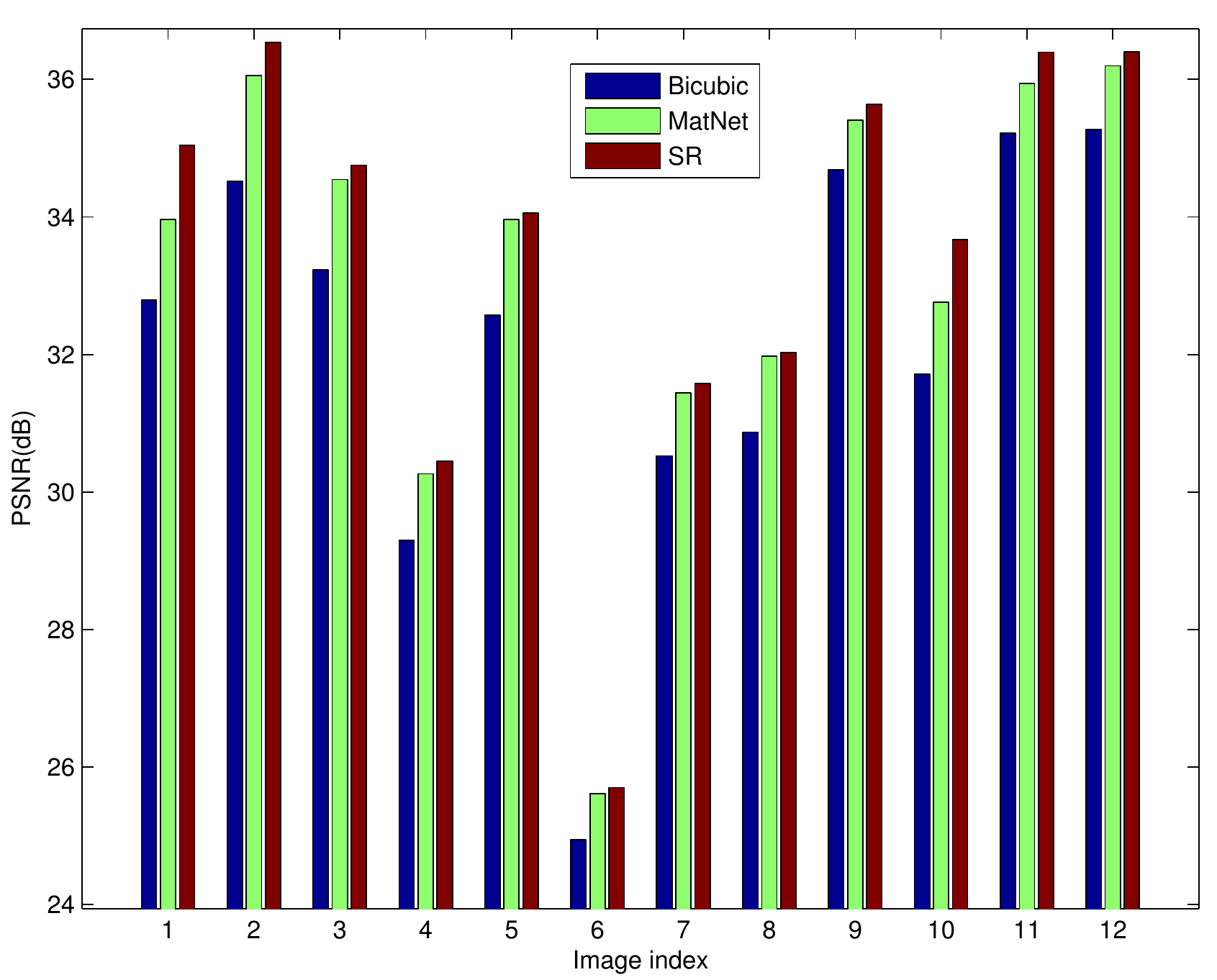}}
  \caption{Super resolution results comparison. The images are indexed from left to right, from top to bottom. }\label{fig:srtestpsnrcomp}
\end{figure}

Fig. \ref{fig:srimages} shows the results on two testing images. Multimodal MatNet has comparable performance as SR, the state-of-the-art super resolution method, evaluated by PSNR: for Lena image, multimodal MatNet, SR and bicubic interpolation achieved PSNR 33.966dB, 35.037dB and 32.795dB respectively; for kingfisher image, they had PSNR 36.056dB, 36.541dB and 34.518dB respectively. We applied to a number of images of similar size ($256\times256$) and we observed similar scenario. Fig. \ref{fig:srtestpsnrcomp} (a) shows the all the test images, including the two in Fig. \ref{fig:srimages}, and PSNR's obtained by different methods is shown in Fig. \ref{fig:srtestpsnrcomp} (b). MatNet is very close to SR in terms of PSNR, especially for image 5 and 8.

\section{Discussion}\label{sec:discussion}
We proposed a matrix neural network (MatNet) in this paper, which takes matrices input directly without vectorisation. The most prominent advantage of MatNet over the traditional vector based neural works is that it reduces the complexity of the optimisation problem drastically, while manages to obtain comparable performance as the state-of-the-art methods. This has been demonstrated in applications of MNIST handwritten digits classification and image super resolution.

As we mentioned several times in the text, MatNet was not specially optimised for the tasks we showed in experiment section. There is a lot of potentials for further improvement. Many techniques used for deep networks can be readily applied to MatNet with appropriate adaptation, e.g. reLU activation function, max-pooling, etc., which certainly become our future research.

% ----------------------------------------------------------------
%\bibliographystyle{chicago}
\bibliographystyle{plain}
\bibliography{refdir/workingref}

%-----------------------------------------------------------------
% Appendix
\section{Appendix}
\section{Backpropagation Algorithm for Regression} We will work out the derivative formulas for all the parameters $\Theta = \{U^{(l)}, V^{(l)}, B^{(l)}\}^L_{l=1}$.
\begin{comment}
From \eqref{eq:MatNet-4}, we can see
\begin{align*}
\frac{\partial L}{\partial U^{(L)}_{ij}} = - \sum^N_{n=1}\text{tr}\left[(Y_n -  X_n^{(L+1)})^T \frac{\partial X_n^{(L+1)}}{\partial U^{(L)}_{ij}}\right] \cdot \end{align*}

Let us consider the element $X^{(l+1)}_{n:ij}$ of $X^{(l+1)}_n$, which is defined by \begin{align*} X^{(l+1)}_{n:ij} = \sigma (\mathbf u^l_i X^{(l)}_n \mathbf v^{lT}_j + b^{(l)}_{ij}) \end{align*} Hence \begin{align*} \frac{\partial X^{(l+1)}_{n:ij}}{\partial u^l_i} = \sigma '(\mathbf u^l_i X^{(l)}_n \mathbf v^{lT}_j + b^{(l)}_{ij})  X^{(l)}_n \mathbf v^{lT}_j \ \ \ \ \text{and} \ \ \ \  \frac{\partial X^{(l+1)}_{n:ij}}{\partial u^l_{i'}} = 0 (i'\not=i) \end{align*} Thus we have \begin{align*} \frac{\partial X^{(l+1)}_{n:ij}}{\partial U^{(l)}}  = \mathbf 1_i \sigma '(\mathbf u^l_i X^{(l)}_n \mathbf v^{lT}_j + b^{(l)}_{ij}) \mathbf v^{l}_j X^{(l)T}_n \end{align*} where $\mathbf 1_i$ is a vector of zeros elements except the $i$-th component being 1.
\end{comment}
We use the following useful formulas
\begin{align*}
&\text{vec}(A X B) = (B^T\otimes A)\text{vec}(X), \\
&\frac{\partial AXB^T}{\partial X} := \frac{\partial \text{vec}(AXB^T)}{\partial \text{vec}(X)} = B\otimes A,
\end{align*}
where $\text{vec}(M)$ transforms a matrix into a column vector along columns of the matrix and $\otimes$ is the Kronecker product operator. Also we will use $\odot$ to denote the elementwise product of two vectors or two matrices. In the following derivative formula for matrix valued functions, we use the tradition $\frac{\partial A}{\partial B} = [\frac{A_{ij}}{\partial B_{kl}}]_{(ij, kl)} \in \mathbb{R}^{(I\times J) \times (K\times L)}$ for matrix variables $A\in\mathbb{R}^{I\times J}$ and $B\in\mathbb{R}^{K\times L}$.

From \eqref{eq:MatNet-2}, we can see that, for all $l=1,2,...,L$ \[ X^{(l+1)}_n = \sigma(N^{(l)}_n),
\]
where $n$ refers to the $n$-th item corresponding to the training dataset.

We are interested in the derivative of the regression loss function \eqref{eq:MatNet-8} with respect to $N^{(l)}_n$. $L$ is the function of $N^{(l)}_n$ via its intermediate variable $N^{(l+1)}_n$. Hence the chain rule gives \begin{align} \text{vec}(\frac{\partial L}{\partial N^{(l)}_n} )^T =  \text{vec}(\frac{\partial L}{\partial N^{(l+1)}_n} )^T \frac{\partial N^{(l+1)}_n}{\partial N^{(l)}_n}. \label{eq:MatNetmodel0} \end{align}

Note that
\begin{align*}
N^{(l+1)}_n = U^{(l+1)} X^{(l+1)} V^{(l+1)T} + B^{(l+1)} \\= U^{(l+1)} \sigma (N^{(l)}_n) V^{(l+1)T} + B^{(l+1)}
\end{align*}
As the sigmoid function $\sigma$ is applied elementwise to the matrix, it is easy to show that \begin{align} \frac{\partial N^{(l+1)}_n}{\partial \sigma(N^{(l)}_n)} &= \frac{\partial \text{vec}\left(N^{(l+1)}_n\right)}{\partial \text{vec}\left(\sigma(N^{(l)}_n)\right)} = V^{(l+1)} \otimes U^{(l+1)}. \label{eq:MatNetmodel1} \end{align}

A direct calculation leads to
\begin{align}
\frac{\partial \sigma(N^{(l)}_n)}{\partial N^{(l)}_n} &=  \text{diag}( \text{vec}(\sigma '(N^{(l)}) )). \label{eq:MatNetmodel2} \end{align}

Taking \eqref{eq:MatNetmodel1} and \eqref{eq:MatNetmodel2} into \eqref{eq:MatNetmodel0} gives, with a transpose, \begin{align*} \text{vec}(\frac{\partial L}{\partial N^{(l)}_n} ) &= \resizebox{0.4\textwidth}{!}{$\text{diag}( \text{vec}(\sigma '(N^{(l)}_n) )) (V^{(l+1)T} \otimes U^{(l+1)T}) \text{vec}(\frac{\partial L}{\partial N^{(l+1)}_n} )$}\\ &=\text{diag}( \text{vec}(\sigma '(N^{(l)}_n) )) \text{vec}\left(U^{(l+1)T} \frac{\partial L}{\partial N^{(l+1)}_n}V^{(l+1)}\right)\\ &=\text{vec}(\sigma '(N^{(l)}_n) )\text{vec}\left(U^{(l+1)T} \frac{\partial L}{\partial N^{(l+1)}_n}V^{(l+1)}\right) \\ &= \text{vec}\left(\sigma '(N^{(l)}_n) \odot \left(U^{(l+1)T} \frac{\partial L}{\partial N^{(l+1)}_n}V^{(l+1)}\right) \right). \end{align*}  Finally we have proved that \begin{align} \frac{\partial L}{\partial N^{(l)}_n} = \left(U^{(l+1)T} \frac{\partial L}{\partial N^{(l+1)}_n}V^{(l+1)}\right) \odot \sigma '(N^{(l)}_n).\label{eq:MatNetmodel3}
\end{align}

From \eqref{eq:MatNet-8} we have \begin{align} \frac{\partial L}{\partial N^{(L)}_n} = (\sigma(N^{(L)}_n) - Y_n) \odot \sigma '(N^{(L)}_n).\label{eq:MatNetmodel4}
\end{align}

Hence both \eqref{eq:MatNetmodel3} and \eqref{eq:MatNetmodel4} jointly define the backpropagation algorithm.  Let us denote $\delta^{(l)}_n = \frac{\partial L}{\partial N^{(l)}_n} $.

Now consider the derivatives with respect to parameters. Take $U^{(l)}$ as an example:
\begin{align*}
\text{vec}\left(\frac{\partial L}{\partial U^{(l)}}\right)^T &= \sum^N_{n=1} \text{vec}\left(\frac{\partial L}{\partial N^{(l)}_n}\right)^T \frac{\partial N^{(l)}_n}{\partial U^{(l)}}\\ &=\sum^N_{n=1} \text{vec}\left(\frac{\partial L}{\partial N^{(l)}_n}\right)^T \left( V^{(l)}X^{(l)T}_n\otimes \mathbf I_{I_{l+1}}\right).
\end{align*}
This gives
\begin{align}
\frac{\partial L}{\partial U^{(l)}} = \sum^N_{n=1} \frac{\partial L}{\partial N^{(l)}_n} V^{(l)} X^{(l)T}_n =  \sum^N_{n=1} \delta^{(l)}_n V^{(l)} X^{(l)T}_n. \label{eq:mnnetmodel5} \end{align} Similarly \begin{align}
&\frac{\partial L}{\partial V^{(l)}} =   \sum^N_{n=1} \delta^{(l)T}_n U^{(l)} X^{(l)}_n. \label{eq:mnnetmodel6}\\
&\frac{\partial L}{\partial B^{(l)}} = \sum^N_{n=1} \delta^{(l)}_n\label{eq:mnnetmodel7}
\end{align}

Then we have the following algorithm, for $l = L-1, ..., 1$, \begin{align} &\delta^{(L)}_n = (\sigma(N^{(L)}_n) - Y_n) \odot \sigma '(N^{(L)}_n). \label{eq:mnnetmodel8}\\ &\frac{\partial L}{\partial U^{(l)}} = \sum^N_{n=1} \delta^{(l)}_n V^{(l)} X^{(l)T}_n\label{eq:mnnetmodel9}\\ &\frac{\partial L}{\partial V^{(l)}} = \sum^N_{n=1}\delta^{(l)T}_n U^{(l)} X^{(l)}_n\label{eq:mnnet-20}\\ &\frac{\partial L}{\partial B^{(l)}} = \sum^N_{n=1} \delta^{(l)}_n\label{eq:mnnet-21}\\
&\delta^{(l)}_n = \left(U^{(l+1)T} \delta^{(l+1)}_nV^{(l+1)}\right) \odot \sigma '(N^{(l)}_n) \label{eq:mnnet-22}\\ &\sigma '(N^{(l)}_n) = \sigma(N^{(l)}_n)\cdot (1 - \sigma(N^{(l)}_n)) = X^{(l+1)}_n \cdot (1 - X^{(l+1)}_n) \label{eq:mnnet-23} \end{align}  where $\sigma(N^{(l)}_n) = X^{(l+1)}_n$ is actually the output of layer $l+1$.

\section{Backpropagation Algorithm for Classification}
The only difference between regression and classification mnnet is in the last layer where the output at layer $L+1$ is a vector of dimension $K$. That is the connection between this output layer and layer $L$ is between a vector and the matrix variable $X^{(L)}$ of dimensions $I_L\times J_L$.

According to \eqref{eq:MatNet-7}, we have the following two cases for calculating $\frac{\partial o_{nk}}{\partial N^{(L)}_{nk'}}$:
\begin{itemize}
\item[Case 1:]  $k=k'$. Then
\begin{align*}
\frac{\partial o_{nk}}{\partial N^{(L)}_{nk}} &= \resizebox{0.3\textwidth}{!}{$\frac{\left(\sum^K_{k'=1}\exp(N^{(L)}_{nk'})\right)\exp(N^{(L)}_{nk}) - \exp(N^{(L)}_{nk})\exp(N^{(L)}_{nk})}{\left(\sum^K_{k'=1}\exp(N^{(L)}_{nk'})\right)^2}$}\\
&=o_{nk}(1- o_{nk})
\end{align*}
\item[Case 2:]  $k\not=k'$. Then
\begin{align*}
\frac{\partial o_{nk}}{\partial N^{(L)}_{nk'}} &= \frac{ - \exp(N^{(L)}_{nk})\exp(N^{(L)}_{nk'})}{\left(\sum^K_{k'=1}\exp(N^{(L)}_{nk'})\right)^2} = -o_{nk} o_{nk'} \end{align*} \end{itemize}

Combining the above cases results in
\begin{align*} \delta^{(L)}_{nk} & = \frac{\partial L}{\partial N^{(L)}_{nk}} = -\frac{\partial}{\partial N^{(L)}_{nk}}\sum^K_{k'=1} t_{nk'}\log o_{nk'} \\ &= - t_{nk} \frac1{o_{nk}}o_{nk}(1-o_{nk}) +\sum^K_{k'\not=k}t_{nk'}\frac1{o_{nk'}}o_{nk}o_{nk'}\\
&=o_{nk} - t_{nk}.
\end{align*}

For our convenience, denote
\[
\delta^{(L)} = O_K - T_K = [o_{nk} - t_{nk}]_{nk} \in \mathbb{R}^{N\times K}.
\]

Finally we want to calculate $\delta^{(L-1)}_n = \frac{\partial L}{\partial N^{(L-1)}_n}$ where $N^{(L-1)}_n$ is a matrix, i.e., the output before sigmoid in layer $L$. In other lower layers, the formulas will be the same as the regression case. From  \eqref{eq:MatNet-9},  we have, noting that $N^{(L)}_{nk} = \text{vec}(\sigma(N^{(L-1)}_n)^T \overline{\mathbf u}_k + tb_k$ (\eqref{eq:MatNet-6} and \eqref{eq:MatNet-7}),
\begin{align*}
\text{vec}\left(\frac{\partial L}{\partial N^{(L-1)}_n }\right)& = \sum^K_{k=1} \frac{\partial L}{\partial N^{(L)}_{nk} } \frac{\partial N^{(L)}_{nk}}{\partial N^{(L-1)}_n } \\
= \sum^K_{k=1} \delta^{(L)}_{nk} \text{diag}(\text{vec}(\sigma'(N^{(L-1)}_n))) \overline{\mathbf u}_k.
\end{align*}
For each $\overline{\mathbf u}_{k}$, we convert it into a matrix, denoted by $\overline{U}_k$, according to the position of elements $X^{(L)}_n$, and formulate a third-order tensor $\mathcal{U}$ such that $\mathcal{U}(:,:,k) = \overline{U}_k$.  Then
\begin{align} \delta^{(L-1)}_n = \frac{\partial L}{\partial N^{(L-1)}_n }& =  \sum^K_{k=1} \delta^{(L)}_{nk} (\sigma'(N^{(L-1)}_n) \odot \overline{U}_k)  \notag \\ &= \sigma'(N^{(L-1)}_n) \odot (\mathcal{U}\overline{\times}_3 \delta^{(L)}_n) \label{eq:mnnet-24}
\end{align}

Again, according to both \eqref{eq:MatNet-6} and \eqref{eq:MatNet-7}, it is easy to see that
\begin{align*} \frac{\partial o_{nk}}{\partial \overline{\mathbf u}_{k'}} &= \frac{-\exp(N^{(L)}_{nk}) \exp((N^{(L)}_{nk'}) \text{vec}(X^{(L)}_{n})}{\left(\sum^K_{k'=1}\exp(N^{(L)}_{nk'})\right)^2} \\
&= - o_{nk} o_{nk'} \text{vec}(X^{(L)}_n).
\end{align*}
The second case of $k=k'$ is actually
\begin{align*} \frac{\partial o_{nk}}{\partial \overline{\mathbf u}_{k}} & = \resizebox{0.45\textwidth}{!}{$\frac{ \left(\sum^K_{k'=1}\exp(N^{(L)}_{nk'})\right) \exp(N^{(L)}_{nk}) \text{vec}(X^{(L)}_n) -\exp(N^{(L)}_{nk}) \exp((N^{(L)}_{nk}) \text{vec}(X^{(L)}_{n})}{\left(\sum^K_{k'=1}\exp(N^{(L)}_{nk'})\right)^2}$}  \\
& = o_{nk} (1 - o_{nk}) \text{vec}(X^{(L)}_n)
\end{align*}
Hence, for each $k = 1, 2, ..., K$,
\begin{align} \frac{\partial L}{\partial \overline{\mathbf u}_{k}} &= -\sum^N_{n=1}\sum^K_{k'=1} t_{nk} \frac1{o_{nk'}} \frac{\partial o_{nk'}}{\partial \overline{\mathbf u}_{k}}\notag\\
&=\resizebox{0.45\textwidth}{!}{$-\sum^N_{n=1} \left[\sum^K_{k'\not=k} t_{nk'} \frac1{o_{nk'}} (-o_{nk'})o_{nk} \text{vec}(X^{(L)}_n) + t_{nk} \frac1{o_{nk}}o_{nk}(1-o_{nk})\text{vec}(X^{(L)}_n)\right]$}\notag\\
&=-\sum^N_{n=1}\left[-\left(\sum^K_{k'\not=k}t_{nk'}\right) o_{nk} + t_{nk}(1-o_{nk})\right]\text{vec}(X^{(L)}_n)\notag\\
&=-\sum^N_{n=1}\left[-\left(1-t_{nk}\right) o_{nk} + t_{nk}(1-o_{nk})\right]\text{vec}(X^{(L)}_n)\notag\\
&=\sum^N_{n=1}(o_{nk} - t_{nk}) \text{vec}(X^{(L)}_n) \notag
\end{align}

If we formulate a matrix $\overline{U} = [\overline{\mathbf u}_1, \overline{\mathbf u}_2, ..., \overline{\mathbf u}_K]$, then \begin{align} \frac{\partial L}{\partial \overline{U}} &= \sum^N_{n=1}\text{vec}(X^{L)}_n) [o_{n1} - t_{n1}, o_{n2} - t_{n2}, ..., o_{nK} - t_{nK}]\notag\\ & = \mathbf X^{(L)} \delta^{(L)}\label{eq:mnnet-25} \end{align} where $\mathbf X^{(L)} = [\text{vec}(X^{L)}_1), \text{vec}(X^{L)}_2), ..., \text{vec}(X^{L)}_N)]\in\mathbb{R}^{(I_L\times J_L) \times N}$.

Similar to $\frac{\partial o_{nk}}{\partial N^{(L)}_{nk'}}$, we have \begin{align} \frac{\partial o_{nk}}{\partial tb_k} = o_{nk}(1-o_{nk}) \ \ \text{and}\ \ \  \frac{\partial o_{nk}}{\partial tb_k} = -o_{nk}o_{nk'}  \ \ \ (k\not=k'). \label{eq:mnnet-26} \end{align}

So it is easy to show
\[
\frac{\partial L}{\partial tb_k}  = \sum^N_{n=1} (o_{nk} - t_{nk}), \ \ \ , \text{that is} \ \ \ \ \frac{\partial L}{\partial tb}  = \text{sum}(O_K - T_K).
\]

The entire backpropagation is to combine \eqref{eq:mnnetmodel9} to \eqref{eq:mnnet-23},  and \eqref{eq:mnnet-24} to \eqref{eq:mnnet-26}.

\section{Sparsity}
We repeat the sparsity penalty $R_l$ here.
\begin{align}
R_l = \text{sum}\left( \rho \log\frac{\rho}{\overline{\rho}^{(l)}} + (1-\rho)\log\frac{1-\rho}{1-\overline{\rho}^{(l)}}\right)
\end{align}
where $\text{sum}(M)$ means the sum of all the elements of matrix $M$, and $\log$ and $/$ are applied to matrix elementwise.

If we applied the sparsity constraints on all the layers excepts for input and output layers, the objective function (of regression) defined in \eqref{eq:MatNet-4} can be sparsely regularised as \begin{align} L' = L + \beta\sum^{L}_{l=2} R_l  = \sum^N_{n=1}\frac12 \| Y_n - X^{(L+1)}_n\|^2_F  + \beta\sum^{L}_{l=2} R_l. \label{eq:mnnet-L} \end{align} Then, by noting that $R_j$ ($j<l+1$) is irrelevant to $N^{(l)}_n$,
\begin{align*} &\frac{\partial L'}{\partial N^{(l)}_n} = \frac{\partial}{\partial N^{(l)}_n} (L+\beta \sum^L_{j>l+1}R_j) + \beta\frac{\partial }{\partial N^{(l)}_n} R_{l+1}\\ &= \frac{\partial L'}{\partial N^{(l)}_n} +  \beta\frac{\partial }{\partial N^{(l)}_n} R_{l+1}\\ &= \frac{\partial L'}{\partial N^{(l+1)}_n} \frac{\partial N^{(l+1)}_n}{\partial N^{(l)}_n}+  \beta\frac{\partial }{\partial N^{(l)}_n} R_{l+1}\\ &=  \left[U^{(l+1)T} \frac{\partial L'}{\partial N^{(l+1)}_n} V^{(l+1)}\right]  \odot \sigma'(N^{(l)}_n) +  \beta\frac{\partial }{\partial N^{(l)}_n} R_{l+1} \end{align*}
By using the similar technique, we can prove that \[ \frac{\partial }{\partial N^{(l)}_n} R_{l+1} = \left[-\frac{\rho}{\overline{\rho}^{(l+1)}} + \frac{1-\rho}{1-\overline{\rho}^{(l+1)}}\right]\odot \sigma'(N^{(l)}_n) \] Hence the backpropagation defined in \eqref{eq:MatNetmodel3} can be re-defined as \[ \delta^{'(l)}_n = \resizebox{0.25\textwidth}{!}{$\left[U^{(l+1)T} \delta^{'(l+1)}_n V^{(l+1)} + \beta \left(-\frac{\rho}{\overline{\rho}^{(l)}} + \frac{1-\rho}{1-\overline{\rho}^{(l)}}\right)\right]$} \odot \sigma'(N^{(l)}_n) \]

The above can be easily implemented into BP scheme as explained in previous section.

%%%% FOR multimodal mnnet
\section{BP Algorithm for Multimodal MatNet Autoencoder}
To train the multimodal MatNet autoencoder, we need to work out the derivatives of $L$ with respect to all the parameters. First, we define the derivative of $L$ with respect to the output layer variables \[ \delta^2_{ij} = \widehat{X}^j_{i} - X^j_i.
\]
Now we back-propagate these derivatives from output layer to the hidden layer according to the network structure and define \[ \delta^1_i = \sum^D_{j=1} S_j (\delta^2_{ij}\odot \sigma '(R_jY_iS^T_j + C_j) ) R^T_j = \sum^D_{j=1}R^T_j (\delta^2_{ij}\odot \sigma'(\widehat{X}^j_i) ) S_j \] Then it is not hard to prove that \begin{align} \frac{\partial L}{\partial R_j} &= \frac1N\sum^N_{i=1} (\delta^2_{ij} \odot \sigma '(\widehat{X}^j_i) ) S_j Y^T_i  \label{eq:mnnetmm-4}\\ \frac{\partial L}{\partial S_j} &= \frac1N\sum^N_{i=1} (\delta^2_{ij} \odot \sigma '(\widehat{X}^j_i) )^T  R_j Y_i \label{eq:mnnetmm-5}\\ \frac{\partial L}{\partial C_j} &= \frac1N\sum^N_{i=1} (\delta^2_{ij} \odot \sigma '(\widehat{X}^j_i) ) \label{eq:mnnetmm-6} \end{align} and \begin{align} \frac{\partial L}{\partial U_j} &= \frac1N\sum^N_{i=1} (\delta^1_{i} \odot \sigma '(Y_i) ) V_j X^T_i \label{eq:mnnetmm-7}\\ \frac{\partial L}{\partial V_j} &= \frac1N\sum^N_{i=1} (\delta^1_{i} \odot \sigma '(Y_i) )^TU_j X_i \label{eq:mnnetmm-8}\\ \frac{\partial L}{\partial B} &= \frac1N\sum^N_{i=1} (\delta^1_{i} \odot \sigma '(Y_i) )\label{eq:mnnetmm-9} \end{align}

The algorithm implementation is straighforward. In the forward sweeping, from the input, we can get all $Y_i$ and $\widehat{X}^j_i$, then in the backward sweep, all the $\delta$'s can be calculated, then all the derivatives can be obtained from the above formula.

\section{Sparsity in Multimodal MatNet Autoencoder}
If we applied the sparsity constraint on the hidden layer, the objective function defined in \eqref{eq:MatNetmm-3} becomes \begin{align} L' = L + \lambda R_y  =  \frac1{2N}\sum^N_{i=1} \sum^D_{j=1}\| \widehat{X}^j_i -  X^j_i\|^2_F  + \beta R_y. \label{eq:mnnetmm-10} \end{align}

As $R_y$ is independent of $R_j, S_j, C_j$, then $\frac{\partial L'}{\partial R_j} = \frac{\partial L}{\partial R_j}$, $\frac{\partial L'}{\partial S_j} = \frac{\partial L}{\partial S_j}$ and $\frac{\partial L'}{\partial C_j} = \frac{\partial L}{\partial C_j}$.   We can prove that
\[
\frac{\partial R_y}{\partial Y_i} = \frac1N \left[-\frac{\rho}{\overline{\rho}} + \frac{1-\rho}{1-\overline{\rho}}\right] :\triangleq \frac1N\delta(\rho).
\]
Then we have
\begin{align}
\frac{\partial L'}{\partial U_j} &= \frac1N\sum^N_{i=1} ((\delta^1_{i} + \beta \delta(\rho)) \odot \sigma '(Y_i) ) V_j X^T_i \label{eq:mnnetmm-11}\\ \frac{\partial L'}{\partial V_j} &= \frac1N\sum^N_{i=1} ((\delta^1_{i} + \beta \delta(\rho))  \odot \sigma '(Y_i) )^TU_j X_i \label{eq:mnnetmm-12}\\ \frac{\partial L'}{\partial B} &= \frac1N\sum^N_{i=1} ((\delta^1_{i} + \beta \delta(\rho)) \odot \sigma '(Y_i) )\label{eq:mnnetmm-13} \end{align}

\end{document}